\documentclass[runningheads]{llncs}
\usepackage[T1]{fontenc}
\usepackage{graphicx}
\usepackage{amsmath}
\usepackage{amssymb}
\usepackage{booktabs}
\usepackage{url}
\usepackage{bbding}
\usepackage[misc]{ifsym}
\usepackage{multirow}
\usepackage[pagebackref,breaklinks,colorlinks]{hyperref}
\usepackage{subcaption}
\usepackage{color}

\usepackage{cite}
\usepackage[capitalize]{cleveref}
\crefname{section}{Sec.}{Secs.}
\Crefname{section}{Section}{Sections}
\Crefname{table}{Table}{Tables}
\crefname{table}{Tab.}{Tabs.}

\begin{document}
\title{Vision-Language Adaptive Mutual Decoder for OOV-STR}
%
%

\author{Jinshui Hu\inst{1,2} \and Chenyu Liu\inst{1,2} \and Qiandong Yan\inst{2} \and Xuyang Zhu\inst{2} \and \\ Jiajia Wu\inst{2} \and Jun Du\inst{1} \and Lirong Dai\inst{1}\textsuperscript{(\Letter)}}
\authorrunning{J. Hu et al.}
%
\institute{University of Science and Technology of China, Hefei, China \\
\email{\{jshu, cyliu7\}@mail.ustc.edu.cn, \{jundu, lrdai\}@ustc.edu.cn} \and
IFLYTEK Research, Hefei, China \\
\email{\{qdyan, xyzhu8, jjwu\}@iflytek.com}\\
}
\maketitle              
\begin{abstract}
   Recent works have shown huge success of deep learning models for common in vocabulary (IV) scene text recognition. However, in real-world scenarios, out-of-vocabulary (OOV) words are of great importance and SOTA recognition models usually perform poorly on OOV settings. Inspired by the intuition that the learned language prior have limited OOV preformence, we design a framework named Vision Language Adaptive Mutual Decoder (VLAMD) to tackle OOV problems partly. VLAMD consists of three main conponents. Firstly, we build an attention based LSTM decoder with two adaptively merged visual-only modules, yields a vision-language balanced main branch. Secondly, we add an auxiliary query based autoregressive transformer decoding head for common visual and language prior representation learning. Finally, we couple these two designs with bidirectional training for more diverse language modeling, and do mutual sequential decoding to get robuster results. Our approach achieved 70.31\% and 59.61\% word accuracy on IV+OOV and OOV settings respectively on Cropped Word Recognition Task of OOV-ST Challenge at ECCV 2022 TiE Workshop, where we got 1st place on both settings.

\end{abstract}

\section{Introduction}
\label{sec:intro}
Scene text recognition (STR) plays an important part on general visual understanding. Despite of the fast developing in computer vision \cite{resnet,fasterrcnn,maskrcnn,vit} and text recognition \cite{hmm,crnn,aster,sar,abinet}, there are still some problems unsolved in real world scenarios, e.g., the OOV problem. \cite{onvoc} reveals that SOTA methdos perform well on images with words within vocabulary but generalize poorly to images with words outside vocabulary. However, in real-world scenarios OOV words are common and of great importance, for example, toponyms, business names, URLs, random strings, etc. Hence, addapting current systems to recognize OOV instances is a crucial next step forward in terms of both research and application.

To help explore this problem, \cite{oovchallenge} firstly propose a new benchmark that can specifically measure performance over OOV instances. Besides, a challenge named ECCV 2022 Challenge on Out of Vocabulary Scene Text Understanding (OOV-ST) is simultaneously come up, which is held at ECCV 2022 Workshop on Text in Everything (TiE). This benchmark combines common scene-text datasets as training set, i.e. Syn90k\cite{syntext}, ICDAR 2015\cite{icdar15}, TextOCR\cite{textocr}, MLT-19\cite{mlt19}, HierText\cite{hiertext}, OpenTextImages\cite{openimagev5text}. As for the final testset, images with text instances will never occur in the above-mentioned training datasets. This benchmark will serve as the 
footstone of OOV research and OCR community, we also build our work on it.

In this paper, we aim to build up an adaptive and unified recognition framwork for both IV and OOV instances. Our motivation is, the recognizer shall be able to make decision on how much visual and lingustic info is used, when handling different inputs and decode steps. Our contributions can be summarized as follows:
\begin{itemize}
	\item We put forward VLAD module, which can adaptively fuse visual or linguistic features at every decode step, leading to a great improvement on OOV instances. 
	\item Besides, we propose a mutual decoding method together with a TransD head and a bidirectional modeling mechanism, which prevent recognizer from overfitting a undirectional language prior and get robuster co-decoding results both in OOV and IV settings.
	\item We carefully evaluate our proposed VLAMD on official OOV-ST Challenge, which achieves 1st places on both OOV and IV+OOV metric, with a word accuracy of 59.61\% and 70.31\% respectively. Experimental results have proved the effectiveness of our method.
\end{itemize}

\section{Related Works}
\label{sec:rela}
With the development of computer vision and deep learnring, performance of STR has been improved significantly. In this section we'll give a brief review of related works for STR and OOV Recogintion.
\subsection{Scene Text Recognition}
In the past ten years, STR methods evolved from HMM based \cite{hmm}, CTC based \cite{crnn}, to Attention based \cite{LAS,aster,sar,preSync,robustscan,abinet}. Limited by conditional independence assumption, both HMM and CTC based methods perform a language-free fragmented prediction, and usually need an external language model (LM) \cite{ctclm}. On the other hand, attention based approaches adopt an autoregressive framework, bringing an implicit LM and a more flexible spatial recognition ability. Moreover, ASTER \cite{aster} combines a rectification module with an attention decoder to tackle irregular STR, SAR \cite{sar} illustrates that a decoder with 2D attention is a strong baseline that achieves SOTA performance on irregular STR, while SATRN \cite{preSync} extends Transformer \cite{atten} to SAR and outperforms most STR methods. Recently, query based parallel decoders are shown comparable preformance on STR, with a dynamic position enhancement branch \cite{robustscan} or an iterative language modeling \cite{abinet}. Thanks to the development and application of these methods, machines are now able to recognize text in common scenes with a level of accuracy that rivals that of humans, marking a significant breakthrough in the field of text recognition.
\subsection{Out-of-vocabulary Text Recognition}
Despite the considerable development in STR, its application in scenarios with a high number of OOV words still may fail, e.g. $\sim$20\% absolute performence drop from IV to OOV setting \cite{onvoc,oovchallenge}.  However, there are few works aim to tackle OOV problems in text recognition. \cite{onvoc} firstly points a quantitative analysis of OOV text recognition out, reveals the week OOV performence of attention based methods, and proposes a joint learning baseline of attentioned based and segmentation based methods. Inspired by the STR framework of non-autoregressive recognition + LM refinement, RobustScanner \cite{robustscan} designs a decoupled position queried module and dynamicly fuses the positional and language hybrid features, yielding a considerable improvements. These two works have relieved OOV troubles from different views of feature enhancing. On the other direction, \cite{openset} proposes context decoupling modules to solve open-set recognition, while \cite{seqclr,siman,readwrite} attempt to study self supervised learning in STR, we point out here that these directions will also benifit OOV researches further.

\begin{figure*}[t!]
	\centering
	\begin{subfigure}{0.8\textwidth}
		\includegraphics[width=9cm]{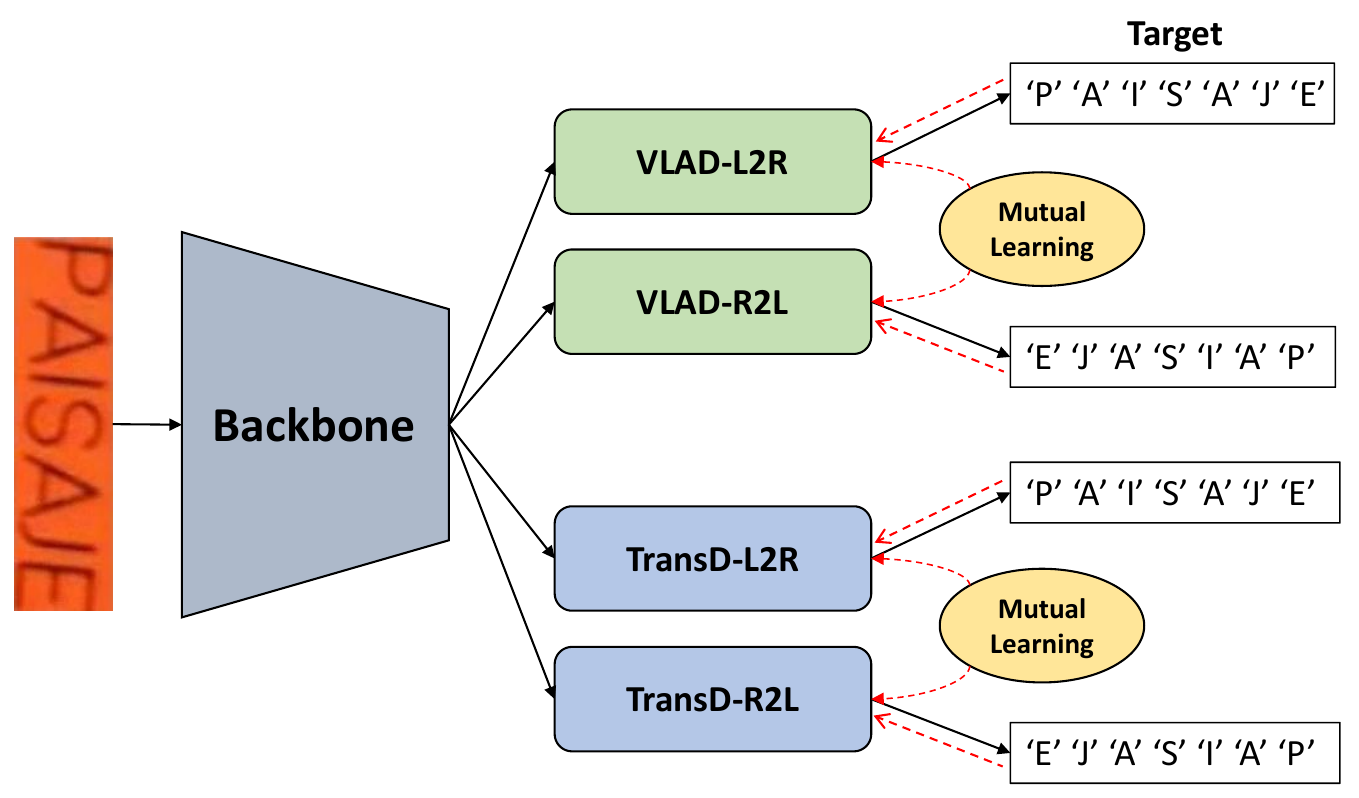}
		\caption{VLAMD Training Framework.}
		\label{framework-a}
	\end{subfigure}
	\\
	\begin{subfigure}{0.8\textwidth}
		\includegraphics[width=9cm]{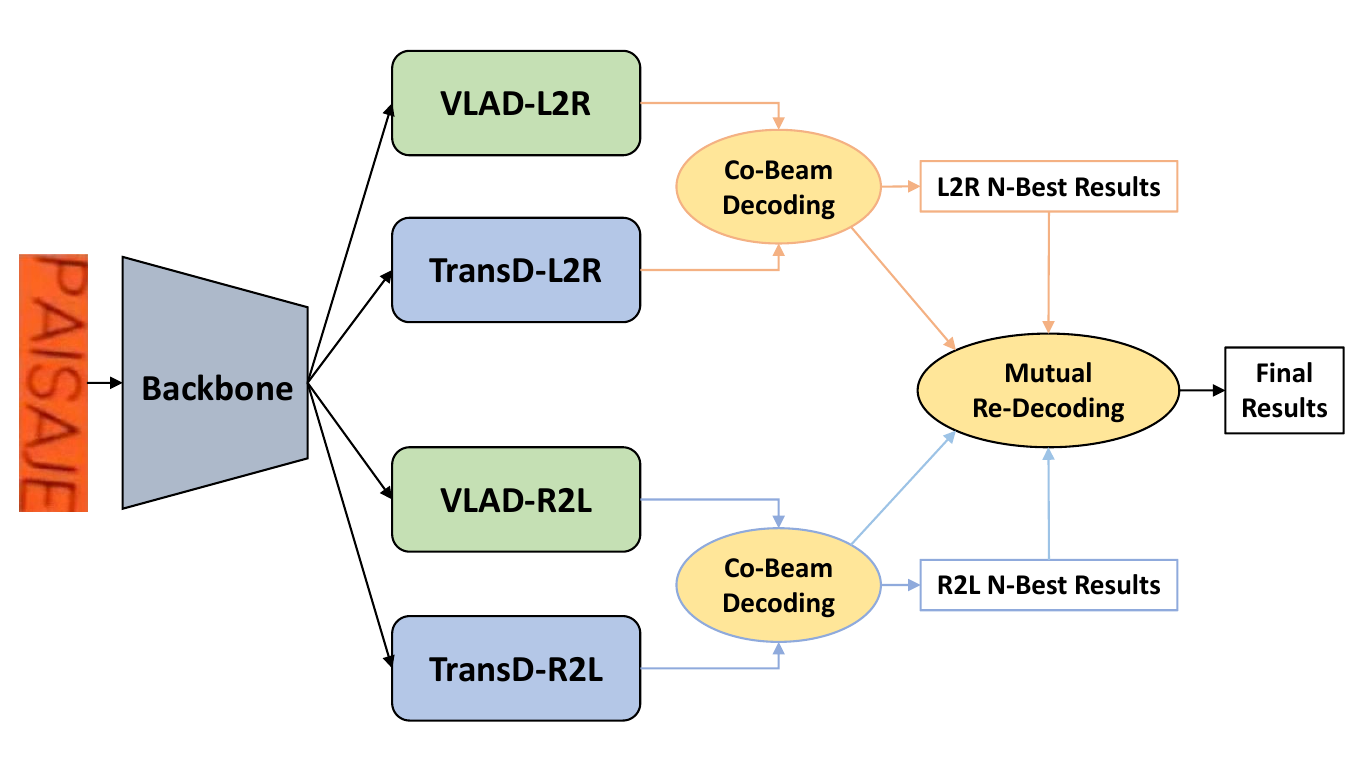}
		\caption{VLAMD Testing Framework.}
		\label{framework-b}
	\end{subfigure}
	\caption{Overall framwork of our proposed VLAMD. VLAMD have three key components: VLAD module, TransD module and the Mutual Decoding strategy. VLAD is designed for dynamicly using visual-lingustic infos, TransD is a common query based transformer decoder, and Mutual Decoding is aiming to get robuster resuls and is used both in training and testing. (a) While training, each of VLAD and TransD will take a bidirectional training head, named L2R and R2L. In addition to commom CE loss, VLAD applies a mutual learning manner through cross KL Divergence to distill extra knowledge. Red dashed lines denote the gradient flows back from losses. (b) During testing, different modules with same decoding direction are combined to form a co-beam decoding, then L2R and R2L N-Best results will do a cross teacher forcing re-decoding, named Mutual Re-Decoding.}
	\label{framework}
\end{figure*}

\section{Approach}
\label{sec:appro}

\subsection{Overview}
The overall training framework can be seen in \cref{framework-a}. Given an image $\mathbf{I}\in \mathbb{R}^{3\times H \times W}$, the backbone will firstly encode it to a downsampled visual contextual feature $\mathbf{F}\in \mathbb{R}^{C\times \frac{H}{4} \times \frac{W}{4}}$. Then, both the VLAD module and the TransD module will decode out two results respectively, i.e., the left to right (L2R) and the right to left (R2L) target strings. These different decode modules together with the backbone are jointly trained, where the loss contains four cross entropy loss guided by the GT target and four mutual KL loss between L2R and R2L sequences, see \cref{loss_all}. 
\subsection{Backbone}
For simplicity, a small CNN Plain-ViT backbone is adopted: 1) we use two Conv blocks with each stride 2 to fast downsample, getting a feature map of size $512\times \frac{H}{4} \times \frac{W}{4}$; 2) we flatten the feature map to $512\times \frac{HW}{16}$, and send it to a standard transformer encoder used in \cite{atten}; 3) Finally, the feature map is reshaped to $512\times \frac{H}{4} \times \frac{W}{4}$ for decoding.
\subsection{VLAMD} %
As mentioned before, our vision-language adaptive mutual decoders for text recognition are made up of three building blocks. One is the key Vision Language Adaptive Decoder (VLAD), one is a query based transformer decoder (TransD), and another is the bidirectional training and the mutual decoding strategy.
\subsubsection{VLAD.}
\begin{figure}[htb]
	\centering
	\includegraphics[width=0.75\textwidth]{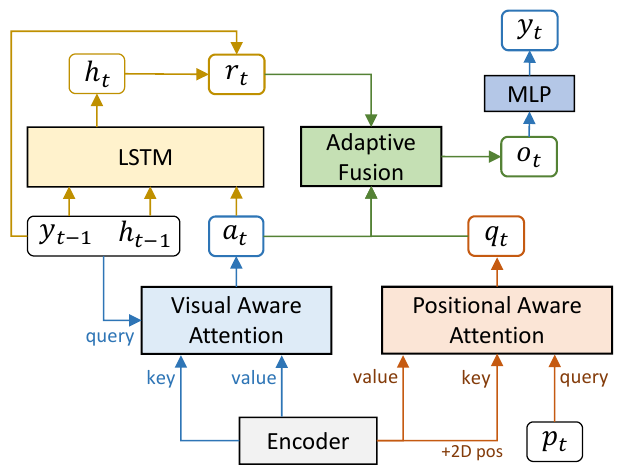}
	\caption{Detailed structure of VLAD. In addition to the main LSTM decoder, VLAD consists of three sub-modules: a) VAA module in light blue, output a visual-only feature $\mathbf{a}_\mathrm{t}$; b) PAA module in light orange red, output a positional aware visual-only feature $\mathbf{q}_\mathrm{t}$; c) Adaptive Fusion module in green, form a key dynamic visual-lingustic fusion strategy.}
	\label{fig:VLAD}
\end{figure}

The main decoding branch is based on LAS \cite{LAS} architecture and coverage attention mechanism \cite{COVERAGE}. Our key insight for vision-language balanced recognition is that, the adaptive choice ability from vision or language shall be a basic component of our system. To achieve this, we additionally design a position query branch named Positional Aware Attention (PAA) module, together with the reusing of the Visual Aware Attention (VAA), to extract visual-only context. Besides, an Adaptive Gated Fusion module is proposed to adaptively fuse the linguistic enhanced feature and the visual-only features. \cref{fig:VLAD} illustrates VLAD in a detailed manner.

Mathematically, during one decoding step $t$, a visual-only context is firstly computed using last step's result $\mathbf{y}_{t-1}$, hidden state $\mathbf{h}_{t-1}$, image feature $\mathbf{F}$:
\begin{equation}
	\mathbf{a}_\mathrm{t} = \mathrm{Attention}([\mathbf{y}_\mathrm{t-1}; \mathbf{h}_\mathrm{t-1}], \mathbf{F}, \mathbf{F}),
	\label{VAA}
\end{equation}
where $\mathrm{Attention}(\mathrm{*query}, \mathrm{*key}, \mathrm{*value})$ denotes a basic attention layer, $\mathbf{a}_\mathrm{t}$ is the obtained visual-only feature. For simplicity, we omit coverage attention here, please refer to \cite{COVERAGE} for detail.

Then, the hidden state $\mathbf{h}_\mathrm{t}$ for LSTM cell and the linguistic enhanced feature $\mathbf{r}_\mathrm{t}$ are updated by
\begin{equation}
	\mathbf{h}_\mathrm{t} = \mathrm{LSTM}([\mathbf{y}_\mathrm{t-1}; \mathbf{a}_\mathrm{t-1}], \mathbf{h}_\mathrm{t-1}),
	\label{lstmcell}
\end{equation}
\begin{equation}
	\mathbf{r}_\mathrm{t} = \mathrm{Concat}(\mathbf{h}_\mathrm{t}, \mathbf{y}_\mathrm{t-1}).
	\label{lstmout}
\end{equation}

For position aware attention module, firstly a position embedding  layer $\mathbf{P}=\left[\mathbf{p}_1, \mathbf{p}_2, \cdots, \mathbf{p}_\mathrm{t^{\prime}}\right]$ is learnt for querying, where the index $t^{\prime}$ denotes the decoding token id. Following \cite{robustscan}, a postion enhanced feature $\mathbf{F^{\prime}}$ is used as the key, and the oringnal $\mathbf{F}$ is used as the value: 
\begin{equation}
	\mathbf{q}_\mathrm{t} = \mathrm{Attention}\left(\mathbf{p}_\mathrm{t}, \mathbf{F^{\prime}}, \mathbf{F}\right),
	\label{paa}
\end{equation}
in which $\mathbf{q}_\mathrm{t}$ is the expected position queried visual-only feature.

Now, for time step $t$, we have got the linguistic enhanced feature $\mathbf{r}_\mathrm{t}$,  the original visual-only context $\mathbf{a}_\mathrm{t}$, the position queried visual-only feature $\mathbf{q}_\mathrm{t}$. Finally, an Adaptive Gated Fusion (AGF) block is proposed to dynamicly merge and balance the visual and linguistic infomations. Specifically, for each decoding step, AGF learns a channel-wise gate $\mathbf{g}_\mathrm{t}$ aotumatically to merge these features:  
\begin{equation}
	\mathbf{g}_\mathrm{t} = \mathrm{Sigmoid}(\mathbf{W}_\mathrm{m}[\mathbf{r}_\mathrm{t}; \mathbf{a}_\mathrm{t}; \mathbf{q}_\mathrm{t}]),
	\label{adap1}
\end{equation}
\begin{equation}
	\mathbf{o}_\mathrm{t} = \mathbf{W}_\mathrm{o}\left[\mathbf{g}_\mathrm{t}\odot \left[\mathbf{r}_\mathrm{t}; \mathbf{a}_\mathrm{t}; \mathbf{q}_\mathrm{t}\right]\right],
	\label{adap2}
\end{equation}
in which $\mathbf{W}_\mathrm{m}$ and $\mathbf{W}_\mathrm{o}$ are learnable FC layers. Then, the final ouput token for step $t$ can be obtained by
\begin{equation}
	\mathbf{y}_\mathrm{t} = \mathrm{Softmax}\left(\mathrm{MLP}\left(\mathbf{o}_\mathrm{t}\right)\right),
	\label{softmax}
\end{equation}
Here $\mathrm{MLP}(\cdot)$ denotes a single layer or a two layer fully connected neural network.
\subsubsection{TransD.}
The second branch in our framework is formed by a naive transformer decoder \cite{atten}. Given the encoded map $\mathbf{F}$, and a learned position enbedding $\mathbf{Q^{\prime}}$ similar to \cref{paa}, the finaly features $\mathbf{O^{\prime}}=[\mathbf{o}_1^{\prime},\cdots, \mathbf{o}_\mathrm{t}^{\prime}]$ and the ouputs $\mathbf{Y^{\prime}}=[\mathbf{y}_1^{\prime},\cdots, \mathbf{y}_\mathrm{t}^{\prime}]$ can be formulated as:
\begin{equation}
	\mathbf{O^{\prime}} = \mathrm{TransD}(\mathbf{Q^{\prime}}, \mathbf{F}),
	\label{TransD1}
\end{equation}
\begin{equation}
	\mathbf{Y^{\prime}} = \mathrm{Softmax}(\mathrm{MLP}(\mathbf{O^{\prime}})),
	\label{TransD2}
\end{equation}
where $\mathrm{TransD}(\cdot)$ represents a stack of tranformer decoder layers, with self attention and cross attention in it. 
\subsubsection{Mutual Decoding.}
As mentioned before, for each branch, we copy it and construst two decoding targets during training and testing. Specifically, for one target string sequence $\mathbf{S}_\mathrm{L2R}=[s_1, s_2, \cdots, s_L]$, we reverse it to $\mathbf{S}_\mathrm{R2L}=[s_L, \cdots, s_2, s_1]$ which will be used as the other target. As shown in \cref{framework-a}, both VLAD and TransD will be added twice and supervised by $\mathbf{S}_\mathrm{L2R}$ and $\mathbf{S}_\mathrm{R2L}$, respectively. We define VLAD's output distribution as $\mathbf{Y}_\mathrm{L2R}$ and $\mathbf{Y}_\mathrm{R2L}$, TransD's output distribution as $\mathbf{Y}^{\prime}_\mathrm{L2R}$ and $\mathbf{Y}^{\prime}_\mathrm{R2L}$, our main loss is:
\begin{align}
	\mathcal{L}_\mathrm{main} = 	&\mathrm{CE}(\mathbf{S}_\mathrm{L2R}, \mathbf{Y}_\mathrm{L2R}) 
	+
	\mathrm{CE}(\mathbf{S}_\mathrm{R2L}, \mathbf{Y}_\mathrm{R2L}) \notag \\
	+
	&\mathrm{CE}(\mathbf{S}_\mathrm{L2R}, \mathbf{Y}^{\prime}_\mathrm{L2R})
	+
	\mathrm{CE}(\mathbf{S}_\mathrm{R2L}, \mathbf{Y}^{\prime}_\mathrm{R2L}),
	\label{loss_main}
\end{align}
where $\mathrm{CE}(\mathrm{*gt, *pred})$ is a standard cross entropy loss.

In order to prevent our model from overfiting an unidirectional single language prior, we propose to apply a bidirectional mutual learning strategy: for the same branch, we make them distill from each other using the L2R and R2L head on the every same token. Hence, two cross Kullback-Leibler Divergence (KLD) loss with stop gradient is used: 
\begin{align}
	\mathcal{L}_\mathrm{mut}(\mathbf{Y}_\mathrm{L2R}, \mathbf{Y}_\mathrm{R2L}) & =  \mathrm{KL}(\mathbf{Y}_\mathrm{L2R}~||~\mathrm{RS}(\mathbf{Y}_\mathrm{R2L})) \notag \\
	& +  \mathrm{KL}(\mathbf{Y}_\mathrm{R2L}~||~\mathrm{RS}(\mathbf{Y}_\mathrm{L2R})),
	\label{kl}
\end{align}
where $\mathrm{KL}(p||q)=\sum\limits_{i=1}^{N}p(x)\log\frac{p(x)}{q(x)}$ denotes a KLD function, $\mathrm{RS}(\cdot)$ denotes a sequence reverse operation followed by a stop gradient layer. The overall loss of our system is:
\begin{align}
	\mathcal{L}_{total} = \mathcal{L}_{main} & + \lambda\cdot\mathcal{L}_\mathrm{mut}(\mathbf{Y}_\mathrm{L2R}, \mathbf{Y}_\mathrm{R2L}) \notag \\ & + \lambda\cdot\mathcal{L}_\mathrm{mut}(\mathbf{Y}^{\prime}_\mathrm{L2R}, \mathbf{Y}^{\prime}_\mathrm{R2L}),
	\label{loss_all}
\end{align}
and $\lambda$ is a hyper parameter.

VLAMD's inference process is shown clearly in \cref{framework-b}. Firstly, the two branch VLAD and TransD will do joint co-beam search process twice, yield a left to right N-Best list and a right to left N-Best list. Then, using a cross teacher forcing scheme, our system applys a mutual decoding method: 
\begin{equation}
	\mathbb{P}(\mathbf{y}_\mathrm{pred}|\mathbf{F}) = \mathbf{Y}_\mathrm{L2R}(\mathbf{y}_\mathrm{pred}) + \mathbf{Y}_\mathrm{R2L}(\mathrm{Reverse}(\mathbf{y}_\mathrm{pred})).
	\label{infer}
\end{equation}
According to \cref{infer}, for a decoding path $\mathbf{y}_\mathrm{pred} = \mathrm{[y_{pred}^1,y_{pred}^2,\cdots,y_{pred}^T}]$ from L2R joint co-beam search result, we will send it to R2L joint co-beam search module and vice versa. We found it efficient to acquire robuster results not only in OOV sets but also in IV+OOV settings.

\section{Experiments}
\label{sec:exper}
\subsection{Datasets}
We evaluate our method on OOV-ST Challenge \cite{oovchallenge} mentioned in \cref{sec:intro}, which contains fine-grained validation and test sets with OOV or IV tags. For training, OOV-ST contains a total of 4.29M real cropped line images collected from several public datasets \cite{icdar15,textocr,mlt19,hiertext,openimagev5text}, and a corpus of 90K common words \cite{syntext} for synthetising new data. Besides, there are 113K cropped lines for validation and 313K cropped lines for testing, respectively.

In our experiments, we do not synthesise any data ourselves. Only a subset of Synth90K \cite{syntext} and SynthText \cite{synth2016} are used for a short pretaining \cite{preSync}, where both are public synthetic datasets using the 90K words. Moreover, we filter out training samples that contain out of dictionary characters, remaining 3.98M real cropped lines for training.

\subsection{Training Details}
For simplicity, all images are resized to 32x100 for both training and testing, and we do not use any data augmentation tricks. We firstly pretrain the backbone using a simple decoder on synthetic data for 4 epochs, and finetune it on 3.98M real image lines using the proposed VLADM for 10 epochs. An ADAM optimizer with multistep lr decay is adopted, and the base learning rate is set to 1e-4, weight decay is set to 1e-5, batch size is set to 128. $\lambda$ in \cref{loss_all} is set to 0.4 after applying a grid search method. 

\begin{table}[tb]
	\centering
	\begin{tabular}{cccc}
		\toprule
		\multirow{2}{*}{~~Rank~~} & \multirow{2}{*}{Method} & \multicolumn{2}{c}{CRW (\%)} \\
		& & OOV & IV+OOV \\
		\midrule
		1 & $\textbf{VLAMD}$ & \textbf{59.61} & \textbf{70.31} \\
		~2~ & ~~~SCATTER~~~  & ~~~59.45~~~ & ~~~69.58~~~ \\
		3 & dat  & 59.03 & 69.90 \\
		4 & MaskOCR  & 58.65 & 69.63\\
		5 & Summer  & 58.06 & 68.77\\
		\bottomrule
	\end{tabular}
	\caption{OOV-ST Challenge Results \cite{oovchallenge} on test set. All results are from the official \cite{oovchallenge}, and we list top 5 methods here accroding to the official OOV CRW metric. Submits from the same affiliation are filted out. Our VLAMD obtains best performance on both OOV and IV+OOV metrics.}
	\label{test-oov}
\end{table}

\begin{table}[tb]
	\centering
	\begin{tabular}{cccc}
		\toprule
		\multirow{2}{*}{Method} & \multicolumn{2}{c}{CRW (\%)} \\
		& OOV & IV+OOV \\
		\midrule
		ABINet\cite{abinet}  & ~~~48.55~~~ & ~~~59.84~~~ \\
		~~~~~UnifiedSTR\cite{strwrong}~~~~~  & 53.96 & 64.97 \\
		SCATTER\cite{scatter}  & 55.38 & 66.68\\
		\midrule
		$\textbf{Ours}$ & \textbf{59.61} & \textbf{70.31} \\
		\bottomrule
	\end{tabular}
	\caption{Comparison with SOTA STR methods on OOV-ST testset. Our method outperforms others by a large margin. (Note that the SCATTER method here is different with \cref{test-oov} SCATTER, SCATTER in \cref{test-oov} is just a team name.)}
	\label{compare-oov}
\end{table}

\subsection{Comparison with State-of-the-Art Methods}

We evaluate VLAMD's performance against state-of-the-art methods using the Correcly Recognized Words (CRW) rate, which is a commonly used metric in speech-to-text recognition (STR) tasks. In \cref{test-oov}, we compare VLAMD's results with those of other participants on the OOV-ST dataset. It is worth noting that most existing methods struggle to balance performance between in-vocabulary (IV) and out-of-vocabulary (OOV) words, while VLAMD achieves the highest CRW rate in both settings, demonstrating its superior performance.

To further illustrate VLAMD's competitiveness, we compare it with other public state-of-the-art STR methods, including ABINet \cite{abinet}, UnifiedSTR \cite{strwrong}, and SCATTER \cite{scatter}, on the OOV-ST dataset \cite{oovchallenge}. These methods are reimplemented using their source code and evaluated their performance using the CRW rate. As shown in \cref{compare-oov}, VLAMD outperforms these methods in both the IV and OOV settings, achieving a CRW rate of 59.61\% and 70.31\%, respectively, outperform the second-best method SCATTER's 55.38\% and 66.78\% by a large margin. These results indicate that VLAMD has achieved state-of-the-art performance in STR tasks, especially in handling and balancing both IV and OOV words.

\begin{figure}[htb]
	\centering
	\includegraphics[width=0.98\textwidth]{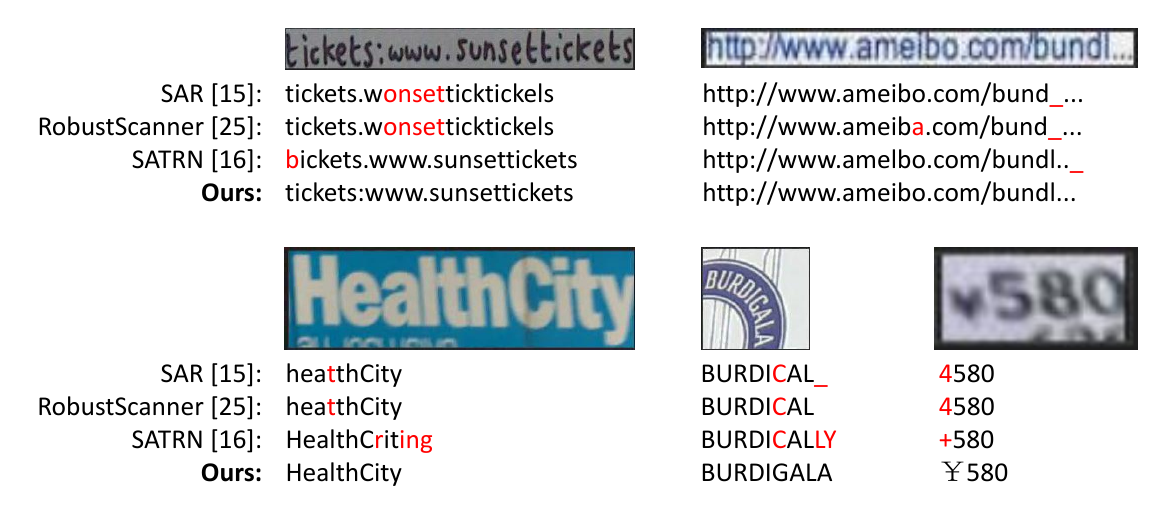}
	\caption{Visualization of recognition results across different methods. The images samples are from OOV-ST OOV validation set \cite{oovchallenge}. For the sake of fairness, all models are trained using the same training data and hyperparameters. Wrong charaters are shown in red color.}
	\label{vis}
\end{figure}

In addition to quantitative evaluations, we also conducted a visual comparison of the outputs generated by different methods. As depicted in \cref{vis}, our analysis indicates that prior techniques suffer from recognition instability and errors when confronted with challenging OOV scenarios, while our proposed VLAMD model exhibits superior and consistent recognition performance. Our conclusion is further bolstered by the compelling evidence presented in the visual comparison, which underscores the effectiveness of our approach.

\begin{table}[tb]
	\centering
	\begin{tabular}{ccccc|cc} 
		\toprule
		~~BS~~ & ~VLAD~ & ~~TD~~ & ~~MT~~ & ~~ES~~ & \multicolumn{2}{@{}c@{}}{\begin{tabular}{cc}
				\multicolumn{2}{c}{CRW (\%)} \\
				~~OOV & ~~~~~~IV+OOV \\
		\end{tabular}}  \\
		\midrule
		\multicolumn{5}{c|}{RobustScanner  \cite{robustscan}} & 60.36 & 71.85 \\
		\midrule
		\checkmark & & & & &   ~~~~~60.42~~~~~ & ~~~~~72.04~~~~~ \\
		\checkmark & \checkmark& & & &   60.82 & 72.35 \\
		\checkmark & \checkmark& \checkmark& & &  61.84 & 73.42 \\
		\checkmark & \checkmark& \checkmark& \checkmark& & \textbf{62.61} & \textbf{73.92} \\
		\checkmark & \checkmark& \checkmark& \checkmark& \checkmark&  \textbf{64.85} &\textbf{75.83} \\
		\bottomrule
	\end{tabular}
	\caption{Abalation study on OOV-ST validation set. Both OOV and OOV+IV metric used in the challenge are evaluated here. BS means our baseline, and VLAD, TD, MT denotes our proposed VLAD, TransD, bidirectional and mutual decoding strategy respectively. ES denotes our 4-ensemble final model submitted to OOV-ST, in which different seeds and heads are used.}
	\label{valid}
\end{table}

\subsection{Ablation Study}
We show the effectiveness of each module in VLAMD on validatioin set, see \cref{valid}. Firstly we reproduce RobustScanner\cite{robustscan} as a strong baseline for comparison, using the public code in \cite{mmocr2021}. And VLAMD's baseline in \cref{valid} is a simple decoder based on \cref{lstmout}, using \cite{LAS,COVERAGE}. Then, each module designed in \cref{sec:appro} is added to our baseline cumulatively for ablation study. As shown, even our baseline can achieve comparable performance with SOTA OOV method, the proposed VLAD, TransD, Mutual Decoding modules are all effective. Finally, our submitted VLAMD in \cref{test-oov} is formed by 4-ensemble models, shown in the last line of \cref{valid}.

\section{Conclusion}
\label{conclusion}
In this paper, we present VLAMD, an adaptive and unified recognition framework for scene text recognition that addresses the OOV problem and balances both IV and OOV scenarios. Our VLAMD dynamically fuses visual and linguistic information and enables bidirectional mutual learning and decoding, resulting in a significant improvement in both OOV and IV STR. Experimental results show the effectiveness of our proposed method, which outperforms SOTA STR methods by a large margin and won the 2022 ECCV OOV-ST Challenge. We hope this work will inspire further research on this important topic.


%
%
%
\bibliographystyle{splncs04}
\bibliography{mybibliography}
%
%
%
%
%
\end{document}